\PassOptionsToPackage{table}{xcolor}
\documentclass[10pt,twocolumn,letterpaper]{article}
\usepackage[pagenumbers]{cvpr}
\usepackage[T1]{fontenc}
\usepackage[utf8]{inputenc}
\usepackage{mathptmx}
\usepackage{microtype}
\usepackage{tabularx}
\usepackage{multirow}
\usepackage{array}
\usepackage{makecell}
\usepackage{colortbl}
\usepackage{float}
\usepackage{algorithm}
\usepackage{algpseudocode}
\usepackage{placeins}
\usepackage{needspace}
\usepackage{listings}
\usepackage{tikz}
\usetikzlibrary{arrows.meta,positioning,calc}
\definecolor{cvprblue}{rgb}{0.21,0.49,0.74}
\definecolor{draftred}{RGB}{148,41,43}
\definecolor{inkblue}{RGB}{36,67,91}
\definecolor{lightblue}{RGB}{237,244,248}
\definecolor{lightgray}{RGB}{246,246,246}
\usepackage[pagebackref,breaklinks,colorlinks,allcolors=cvprblue]{hyperref}

\makeatletter
\renewcommand\paragraph{\@startsection{paragraph}{4}{\z@}{1.0ex plus .2ex minus .2ex}{-0.8em}{\normalfont\normalsize\bfseries}}
\makeatother

\newcommand{\method}{\textsc{SimplePro}}

\definecolor{tablerowgreen}{RGB}{231,247,231}

\definecolor{tablerowgreen}{RGB}{231,247,231}

\newcommand{\papertablestyle}{
  \setlength{\arrayrulewidth}{0.3pt}
  \setlength{\heavyrulewidth}{0.6pt}
  \setlength{\lightrulewidth}{0.35pt}
  \setlength{\cmidrulewidth}{0.3pt}
  \arrayrulecolor{black!60}
}

\lstdefinestyle{prompt}{basicstyle=\ttfamily\footnotesize,breaklines=true,breakatwhitespace=true,columns=fullflexible,keepspaces=true,showstringspaces=false,frame=single,framerule=0.3pt,rulecolor=\color{black!20},backgroundcolor=\color{lightgray},xleftmargin=3pt,xrightmargin=3pt,framesep=4pt}
\lstdefinestyle{paths}{style=prompt,basicstyle=\ttfamily\scriptsize,breakatwhitespace=false}
\title{SimplePro: A Simple Baseline for Proactive Video Understanding}
\author{
Yilin Wang\textsuperscript{1,*}\quad
Xiangxi Zheng\textsuperscript{2,*}\quad
Dongxing Mao\textsuperscript{3}\\[3pt]
Linjie Li\textsuperscript{4}\quad
Zhengyuan Yang\textsuperscript{5}\quad
Alex Jinpeng Wang\textsuperscript{3,\ensuremath{\dagger}}\\[6pt]
{\normalsize\textsuperscript{1}ZJU\quad
\textsuperscript{2}NJU\quad
\textsuperscript{3}CSU\quad
\textsuperscript{4}UW\quad
\textsuperscript{5}Microsoft}\\[3pt]
{\small\textsuperscript{*}Equal contribution.\quad
\textsuperscript{\ensuremath{\dagger}}Corresponding author.}}
\begin{document}
\setcounter{topnumber}{1}
\setcounter{bottomnumber}{1}
\setcounter{totalnumber}{2}
\setcounter{dbltopnumber}{1}
\maketitle
\begin{abstract}
Proactive video understanding requires vision-language models (VLMs) to decide not only \emph{what} to answer as a video unfolds, but also \emph{when} newly arrived evidence warrants a response and when to remain silent.
Although modern VLMs excel at offline long-video understanding, they struggle in proactive streaming settings, leading prior work to introduce task-specific reinforcement learning or auxiliary triggering modules.
We find that explicit reporting rules and structured streaming context substantially improve a frozen VLM's proactive response performance.
We exploit this observation to build \method{}, a training-free baseline for proactive video understanding.
At each decision step, the model combines recent high-resolution frames, bounded historical observations, and prior answer records to determine whether new information warrants a response and, when needed, generate a concise answer.
The system then filters consecutive duplicate answers while retaining the raw outputs as context for subsequent decisions.
Without any parameter updates or auxiliary triggers, \method{} with a 4B backbone consistently outperforms specialized proactive systems across OVO-Bench, StreamingBench-PO, and ProactiveVideoQA.
\method{} shows overall performance gains as backbone size increases from 2B to 32B.
\textbf{We hope this simple yet effective method will serve as a strong baseline for proactive video understanding, and we will release our code and inference pipeline to support future research.}
\end{abstract}

\begin{figure}[!t]
\centering
\includegraphics[width=\columnwidth,trim=7bp 8bp 7bp 9bp,clip]{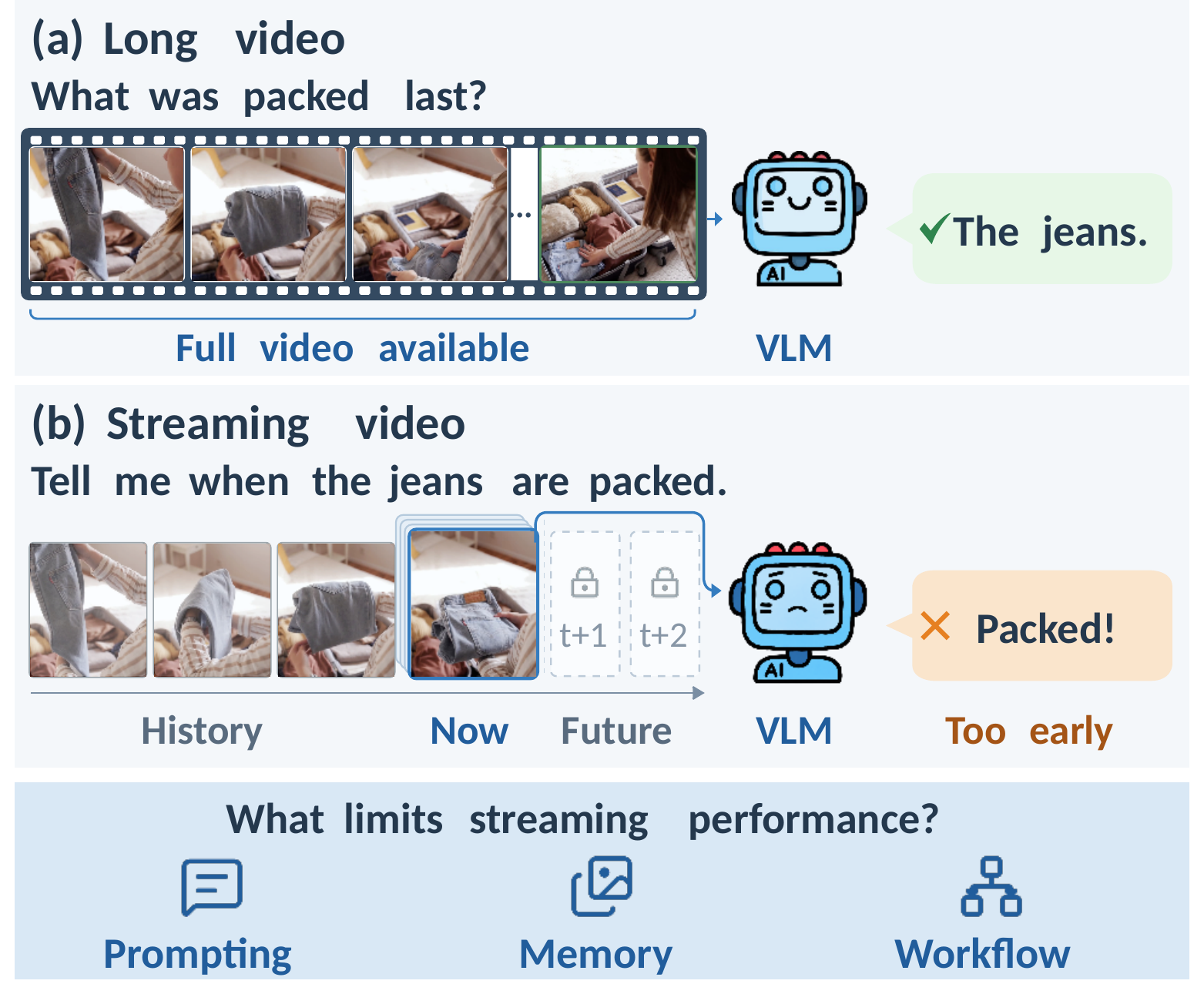}
\caption{Long-video and streaming-video understanding.
(a) Offline models can use the complete video.
(b) Streaming models must decide when to respond from past and current frames.
We study prompting, memory, and workflow for proactive responses.}
\label{fig:teaser}
\end{figure}

\begin{figure*}[!t]
\centering
\includegraphics[width=\textwidth,trim=6bp 4bp 3bp 2bp,clip]{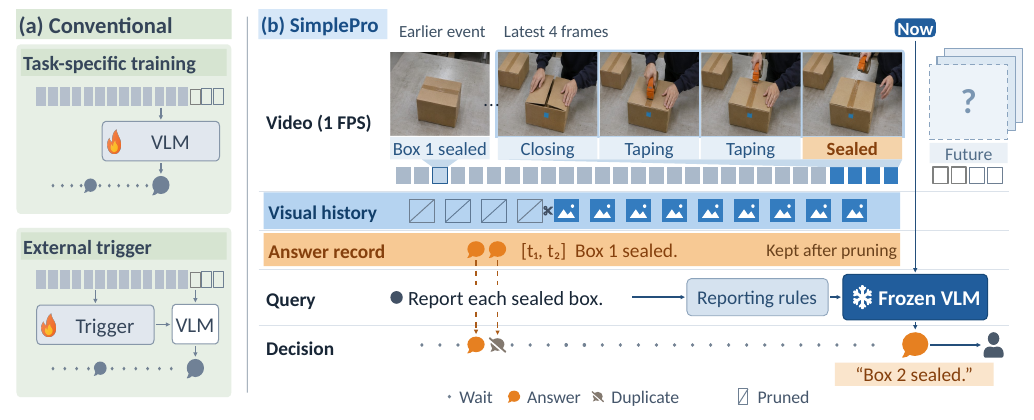}
\caption{Overview of \method{}. Left: Conventional approaches use task-specific training or external triggers. Right: \method{} supplies recent frames, bounded visual history, and prior answers to a frozen VLM. Reporting rules guide joint decisions and answer generation. Observation history is pruned, while consecutive duplicate answers are filtered before delivery. Future frames are unavailable.}
\label{fig:method_scaling}
\vspace{-6pt}
\end{figure*}

\section{Introduction}
\label{sec:intro}

Imagine asking an assistant that watches a live video with you to ``tell me when the water starts boiling'' or ``let me know each time someone enters the room.''
A useful assistant speaks up at the right moment, stays silent otherwise, and does not repeat what it has already said.
This is the goal of proactive video understanding~\cite{niu2025ovo,lin2026streamingbench,wang2025proactivevideoqa}.
Vision-language models (VLMs) now handle offline long videos well~\cite{bai2025qwen25vltechnicalreport,bai2025qwen3,fu2025video,wu2024longvideobench}, yet a streaming assistant faces a different question at every step: does what it has seen so far justify a response (Figure~\ref{fig:teaser})?
Seeing a related object or action does not mean the evidence is sufficient, and an event that continues does not call for another report.

\begin{samepage}
An assistant that answers too early is wrong, one that answers too late misses the moment, and one that answers too often buries the user in duplicates.
\par
\end{samepage}

Existing methods mainly control response timing by adding to the VLM, either through task-specific training or auxiliary triggering mechanisms.
The former use supervised fine-tuning or reinforcement learning to learn when to answer~\cite{wang2025videollm,wang2026mmduet2,li2026videochat3,li2026streampro,tian2026syncpro}. The latter use separate modules to decide when to trigger a response~\cite{qian2025dispider,wang2026streambridge,zheng2026garde,ding2025streammind,azad2026streamready}.
These methods achieve strong results, but the additions have costs: streaming training data, reward design, extra components, and retraining whenever a stronger backbone appears.
Meanwhile, general-purpose VLMs keep improving at perception and instruction following, so it remains unclear how much of the proactive problem a frozen VLM can already solve on its own.
We therefore ask: \textbf{Can explicit reporting rules and structured streaming context enable a frozen VLM to perform proactive video understanding effectively?}

Proactive responses depend on current evidence and prior interactions.
The same scene may call for a different response depending on what has already been reported.
The model therefore needs visual history and prior answers to assess whether the evidence is sufficient and whether another report is needed.
We study how to organize this context and provide explicit reporting rules to guide a frozen VLM.
\par

Guided by this perspective, we introduce \method{}, a training-free baseline for proactive video understanding.
\method{} supplies the model with three types of information: recent high-resolution frames, historical visual observations, and prior answer records.
These provide current evidence, references for event changes, and a record of what the model has already answered, respectively.
The prompt specifies reporting requirements for different questions and allows corrections when new evidence supports them.
In one generation, the model decides whether to respond and produces a concise answer when needed.
The system prunes observation history within a budget while retaining the answer record separately.
A simple delivery filter removes consecutive duplicate answers.
Figure~\ref{fig:method_scaling} illustrates the overall framework.

We evaluate \method{} on three widely used benchmarks: StreamingBench-PO for first-response timing, ProactiveVideoQA for answer quality over time, and OVO-Bench for continued event reporting.
With a frozen 4B backbone, it achieves 54.80\% Acc@2s on StreamingBench-PO, 13.6 points above the strongest compared method, and 52.18 mean PAUC on ProactiveVideoQA, 12.4 points above MMDuet2. On OVO-Bench it reaches 35.23 F1, slightly above SyncPro-RL (34.19), which uses reinforcement learning.
Ablations show that explicit reporting rules improve response performance, while answer records and historical images provide complementary information.
Under a fixed 32K context budget, pruning history yields better results. Reducing historical image resolution cuts input tokens by 36.3\% with little change in F1 score.
Even with reporting rules and answer records, the model still produces consecutive duplicate answers, and delivery filtering further improves results.

Our main contributions are:
\begin{enumerate}[leftmargin=1.4em, label=\arabic*., itemsep=3pt, topsep=3pt, parsep=0pt]
    \item \textbf{A simple but effective baseline.} We introduce \method{}, a training-free baseline for proactive video understanding. It guides a frozen VLM using explicit reporting rules and structured streaming context. With a 4B backbone, it outperforms nearly all compared methods in overall performance across three benchmarks.
    \item \textbf{Evaluation across tasks and model sizes.} We evaluate first-response timing, continued reporting, and answer quality across three benchmarks. We further test frozen backbones from 2B to 32B.
    \item \textbf{Analysis of key design choices.} We evaluate reporting rules, answer records, and visual history, and show the benefits of history compression and duplicate filtering.
\end{enumerate}

\section{Related Work}
\label{sec:related_work}

\paragraph{Video Large Language Models.}
Multimodal large language models align pretrained visual encoders with large language models, acquiring general visual understanding and instruction-following abilities~\cite{alayrac2022flamingo,li2023blip,liu2023visual}, and have been extended to video tasks such as captioning and question answering~\cite{li2025videochat,maaz2024video,lin2024video}.
For long videos, subsequent work improves efficiency and accuracy by compressing visual tokens~\cite{li2024llama,shen2024longvu}, selecting keyframes before inference~\cite{liu2025bolt,tang2025adaptive,yu2025frame,wang2026efficient}, and introducing memory or extending the context length~\cite{song2024moviechat,zhang2024long}.
Recent general-purpose VLMs~\cite{li2024llava,bai2025qwen25vltechnicalreport,bai2025qwen3,wang2025internvl3,zhang2025videollama} continue to set new results on offline benchmarks such as MVBench~\cite{li2024mvbench}, Video-MME~\cite{fu2025video}, and LongVideoBench~\cite{wu2024longvideobench}.
However, these models are designed for the offline setting: the entire video is available before inference, and the model answers once after the user asks.
They therefore do not directly apply to streaming scenarios, where frames arrive continuously.

\paragraph{Reactive Streaming Understanding.}
Streaming video understanding requires a model to process continuously arriving frames in temporal order and to respond to the user based only on what it has observed~\cite{lin2026streamingbench,niu2025ovo,yang2025svbench}.
Existing streaming methods mainly address reactive questions, such as backward tracing and real-time perception, where the model answers immediately after a query based on past or current frames~\cite{niu2025ovo}.
Their central challenge is to retain historical information under limited compute and memory budgets.
Representative approaches maintain external or hierarchical memory~\cite{qian2024streaming,zhang2024flash,huang2025online,zeng2026streamforest,xie2026fluxmem}, compress or retrieve the KV cache~\cite{di2025streaming,yang2025streammem,xu2026streamingvlm}, or prune visual tokens~\cite{wu2024videollm,yao2025timechat,jin2025streamingassistant}.
SimpleStream further shows that an off-the-shelf model given only the most recent frames can match complex memory-based methods~\cite{shen2026simple}.

\paragraph{Proactive Streaming Understanding.}
These reactive tasks follow an ask-then-answer pattern and can be viewed as offline video understanding up to the query time~\cite{wang2025proactivevideoqa}.
Proactive tasks instead require the model to decide on its own when to respond as the video continues, and are evaluated on both answer content and the temporal accuracy of the response sequence~\cite{chen2024videollm,wang2025proactivevideoqa}.
StreamingBench and OVO-Bench evaluate the timing of the first response and the sequence of multiple responses, respectively~\cite{lin2026streamingbench,niu2025ovo}, ProactiveVideoQA measures response quality over time with the PAUC metric~\cite{wang2025proactivevideoqa}, and other benchmarks further cover multimodal inputs, mobile devices, live streaming, and task guidance~\cite{wang2025omnimmi,shi2026river,lu2026phostream,wang2026livibench,zhang2025proactive}.

Existing proactive methods mainly rely on training or additional modules to decide when to respond.
End-to-end methods, represented by VideoLLM-online, predict after each frame whether to stay silent or respond~\cite{chen2024videollm,wang2025videollm,yang2026livestar,xia2026streaming,zhang2025eyes,li2026videochat3}, and recent work further optimizes response timing with reinforcement learning~\cite{wang2026mmduet2,li2026streampro} or introduces ongoing reasoning~\cite{liu2026thinking,guan2026video,tian2026syncpro}.
Decoupled methods instead trigger responses with separate decision, gating, or activation modules~\cite{qian2025dispider,wang2026streambridge,yang2025streamagent,ding2025streammind,zheng2026garde,azad2026streamready}.
These methods typically require additional training data, reward design, or auxiliary components, and often need retraining for a new backbone.
The training-free scheme proposed in OVO-Bench continuously prompts the model to judge whether the target event has occurred~\cite{niu2025ovo}, but models often fail to follow such prompts reliably~\cite{lin2026streamingbench,zheng2026garde}.

In contrast, \method{} introduces no additional trainable parameters or triggering modules.
Instead, it uses explicit reporting rules and structured streaming context so that a single frozen VLM decides whether to respond and generates the answer in one call, distinguishing newly occurring events from ongoing ones.

\section{Method}
\label{sec:method}

\method{} enables a frozen VLM to decide when to respond and what to say as a video unfolds. We first formulate this task as a sequence of decisions based on the observed video and prior interactions (Section~\ref{sec:overview}). We then describe how streaming context and reporting rules guide response generation (Section~\ref{sec:joint-generation}). Finally, we explain how memory is updated and managed as the video progresses (Section~\ref{sec:memory-management}).

\subsection{Proactive Streaming Formulation}
\label{sec:overview}

A user issues a query $q$ at video time $t_0$. The query remains active as new frames arrive, and the system decides when to respond without further user prompts. Depending on the query, the interaction may require a single answer or multiple reports over time. At each decision time $t$, the system uses only the video observed so far and its prior interactions.

We organize the available information into a context $C_t=(F_t,H_t,A_t)$, consisting of recent frames $F_t$, historical observations $H_t$, and an answer record $A_t$. A reporting mode $m$ specifies whether the query calls for a single answer or repeated reports. Given these inputs, the frozen VLM generates a response decision $d_t$ and an optional answer $y_t$ in a single call:
\begin{equation}
(d_t,y_t)=\mathrm{VLM}(q,C_t,m).
\label{eq:simplepro}
\end{equation}
Here, $d_t\in\{\texttt{answer},\texttt{no}\}$. When the model selects \texttt{answer}, it also generates the answer $y_t$. When it selects \texttt{no}, the answer is empty. As new frames arrive, the system updates the context and continues making decisions. Algorithm~\ref{alg:streaming-inference} summarizes this process.

\begingroup
\makeatletter
\renewcommand{\fs@ruled}{
    \def\@fs@cfont{\bfseries}
    \let\@fs@capt\floatc@ruled
    \def\@fs@pre{\hrule height 0.6pt\kern 2pt}
    \def\@fs@mid{\kern 2pt\hrule height 0.35pt\kern 3pt}
    \def\@fs@post{\kern 3pt\hrule height 0.6pt}
    \let\@fs@iftopcapt\iftrue
}
\makeatother
\begin{algorithm}[!t]
\captionsetup{labelsep=colon}
\caption{\method{} Streaming Inference}
\label{alg:streaming-inference}
\fontsize{9}{10.5}\selectfont
\algrenewcommand\algorithmicindent{1.2em}
\algrenewcommand\algorithmicrequire{\textbf{Input:}}
\begin{algorithmic}
\Require Query $q$ and video stream
\State Initialize $H=A=\varnothing$; infer mode $m$\vspace{2pt}
\For{each decision time $t$}
    \Statex \hspace{\algorithmicindent}$\blacktriangleright$ \textbf{Prepare context}
    \State Update recent frames $F_t$
    \State Prune $H$ to meet the context budget
    \State $C_t \gets (F_t,H,A)$\vspace{2pt}
    \Statex \hspace{\algorithmicindent}$\blacktriangleright$ \textbf{Generate and deliver}
    \State $(d_t,y_t) \gets \mathrm{VLM}(q,C_t,m)$
    \If{$d_t=\texttt{answer}$}
        \State Append $(t,y_t)$ to $A$
    \EndIf
    \State Filter and deliver $(d_t,y_t)$\vspace{2pt}
    \Statex \hspace{\algorithmicindent}$\blacktriangleright$ \textbf{Update history}
    \If{an archive update is due}
        \State Apply the history retention policy to $H$
        \State Append the raw interaction to $H$
    \EndIf
\EndFor
\end{algorithmic}
\end{algorithm}
\endgroup

\begin{figure*}[!t]
\centering
\includegraphics[width=\textwidth,trim=19bp 6bp 19bp 7bp,clip]{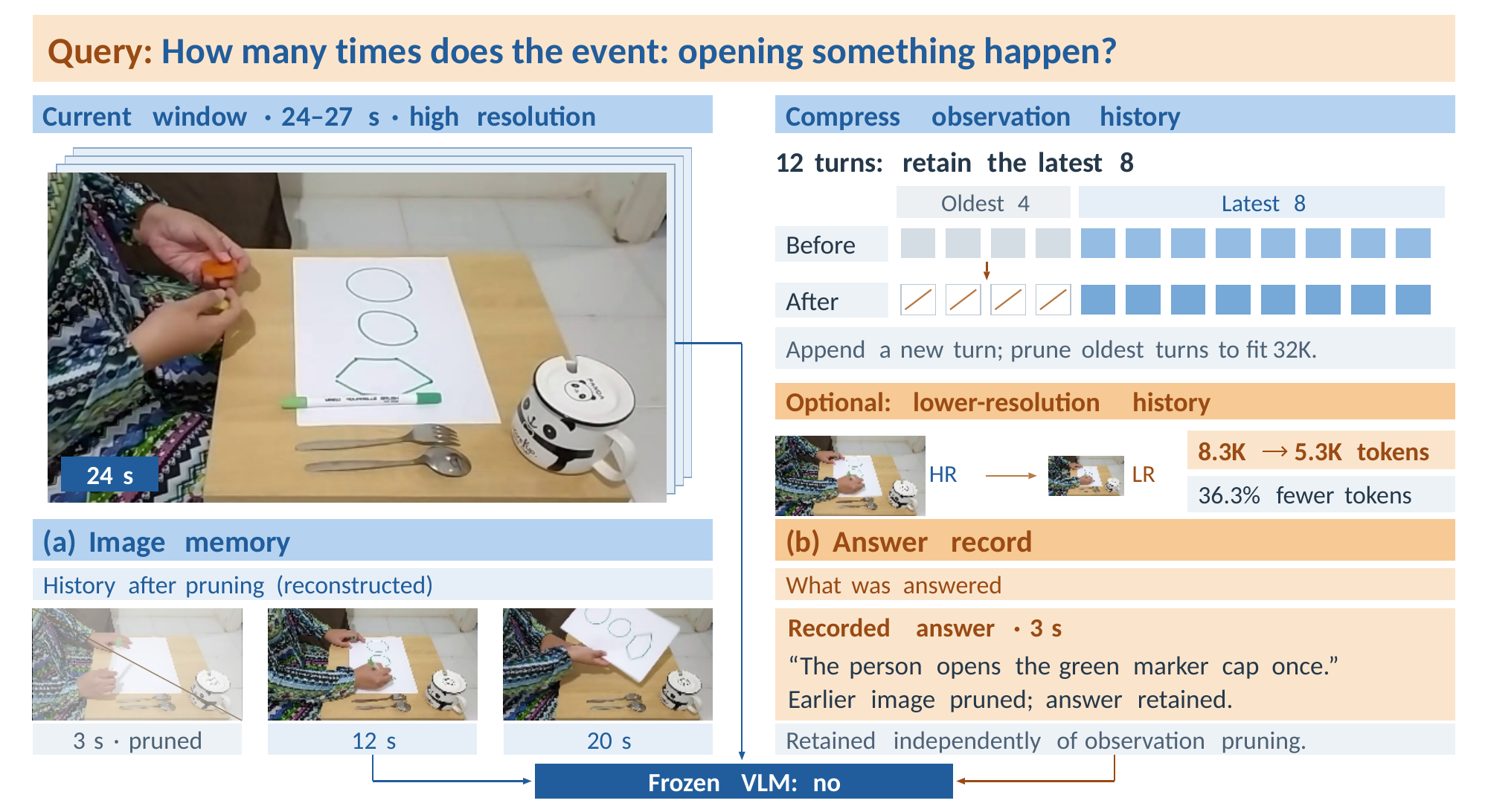}
\caption{Memory organization on the cap-opening example. (a) Historical frames provide earlier visual evidence, with the pruned frame shown for illustration. (b) The answer record retains the opening report at $3\,\mathrm{s}$ after its associated image is pruned. The upper-right panels illustrate bounded history and optional image downsampling.}
\label{fig:component-cases}
\end{figure*}

\subsection{Proactive Response Generation}
\label{sec:joint-generation}
\label{sec:context-construction}

Each response depends on the available information and the user's reporting requirements. We first construct the context for each decision. Reporting rules then guide the model to decide whether to respond and generate an answer.

\paragraph{Context Construction.}
We provide the model with three types of information:
\begin{itemize}[leftmargin=1.2em, itemsep=3pt, topsep=3pt, parsep=0pt]
    \item \textbf{Recent frames ($F_t$)} provide current visual evidence. We sample video at a specified FPS. Each call uses up to four most recent frames at high spatial resolution.
    \item \textbf{Historical observations ($H_t$)} support comparison between the current scene and earlier states. Each history turn contains the original user message, video frames, and raw model reply. The reply includes an answer or a decision to remain silent. Historical frames retain their original high resolution by default.
    \item \textbf{Answer record ($A_t$)} stores the model's previous answers and their video timestamps, helping the model determine whether another report is needed.
\end{itemize}

We arrange retained historical interactions in chronological order.
The current user message contains the query, recent frames, and the answer record with video timestamps.
This lets the model follow the event sequence and relate the current scene to earlier observations and prior answers.

\paragraph{Reporting Rules and Joint Generation.}
Before the first decision, the model determines a reporting mode from the query and the initial video frames. The \texttt{once} mode calls for one answer when sufficient evidence is available. The \texttt{each} mode calls for a report of each requested step or event.

The prompt guides the model to apply these rules using the context above. Visual evidence helps determine whether the question can be answered. Prior answers help determine whether a new response is needed. The model remains silent when evidence is insufficient or an event described in a previous answer is still continuing. New evidence can support a correction to an earlier answer.

For example, a user may ask to be notified when a box is sealed.
Seeing its flaps close is not enough to answer.
The model must wait for  clear visual evidence that sealing is complete.
After the first report, it should remain silent while the same event continues.

\paragraph{Duplicate-Response Filtering.}
Even with instructions to avoid repetition and access to prior answers, the model can still generate consecutive identical answers. We therefore filter consecutive duplicate responses before delivery, while retaining all raw answers as context for subsequent decisions.

\begin{table*}[t]
\centering
\caption{Results on StreamingBench and ProactiveVideoQA. PO uses Acc@2s; ProactiveVideoQA uses $\mathrm{PAUC}$ ($\omega=0.5$). All scores are higher-is-better. Best scores are \textbf{bold}; second best are \underline{underlined}.}
\label{tab:main_results}
\vspace{2pt}
\small
\setlength{\tabcolsep}{5pt}
\renewcommand{\arraystretch}{1.18}
\papertablestyle
\begin{tabularx}{\textwidth}{@{}l c| >{\centering\arraybackslash}p{0.18\textwidth}| *{5}{>{\centering\arraybackslash}X}@{}}
\toprule
\multirow{2}{*}{\textbf{Method}} & \multirow{2}{*}{\textbf{FPS}} & \multicolumn{1}{c|}{\textbf{StreamingBench}} & \multicolumn{5}{c}{\textbf{ProactiveVideoQA (PAUC)}} \\
\cmidrule(lr){3-3} \cmidrule(l){4-8}
& & \textbf{PO (Acc@2s)} & \textbf{WEB} & \textbf{EGO} & \textbf{VAD} & \textbf{TV} & \textbf{Mean} \\
\midrule
\multicolumn{8}{@{}c@{}}{\color{black!55}Streaming models} \\[1pt]
VideoLLM-online~\cite{chen2024videollm} & 2 & 4.00 & 25.90 & 25.00 & 25.00 & 18.30 & 23.55 \\
Dispider~\cite{qian2025dispider} & 1 & 25.30 & -- & -- & -- & -- & -- \\
MMDuet~\cite{wang2025videollm} & -- & 29.44 & 38.90 & 46.00 & 27.40 & 21.10 & 33.35 \\
MMDuet2~\cite{wang2026mmduet2} & 1 & 34.60 & \underline{53.30} & 33.60 & 28.90 & \underline{43.40} & \underline{39.80} \\
Em-Garde~\cite{zheng2026garde} & 2 & 38.00 & 44.30 & \textbf{52.30} & 27.40 & -- & -- \\
VideoChat3~\cite{li2026videochat3} & 1 & -- & 28.40 & 28.10 & 25.10 & -- & -- \\
\midrule
\multicolumn{8}{@{}c@{}}{\color{black!55}Reasoning-enabled models} \\[1pt]
VST~\cite{guan2026video} & 2 & 19.60 & 21.11 & 24.97 & 25.08 & -- & -- \\
ThinkStream~\cite{liu2026thinking} & 2 & 27.20 & 30.19 & 24.47 & 18.17 & -- & -- \\
SyncPro-SFT~\cite{tian2026syncpro} & 1 & 38.80 & 47.83 & 33.89 & 32.14 & -- & -- \\
SyncPro-RL~\cite{tian2026syncpro} & 1 & \underline{41.20} & 48.28 & 33.74 & \underline{32.43} & -- & -- \\
\midrule
\rowcolor{tablerowgreen}
\textbf{\method{}} (Ours) & 1 & \textbf{54.80} & \textbf{54.90} & \underline{52.12} & \textbf{43.86} & \textbf{57.86} & \textbf{52.18} \\
\bottomrule
\end{tabularx}
\end{table*}

\subsection{Memory Management}
\label{sec:memory-management}

Historical interactions accumulate as the video progresses. Longer inputs increase context length and KV cache memory requirements. We therefore limit observation history while maintaining a separate answer record for subsequent decisions (Figure~\ref{fig:component-cases}).

\paragraph{History Updates.}
Every two video seconds, we save the input and raw model reply as one history turn. Each generated answer is also appended to the answer record with its text and video timestamp. A decision to remain silent adds no entry to the answer record.

\paragraph{History Retention.}
We control history length in two ways. First, when the history reaches twelve turns, we retain the latest eight before appending the current turn. Second, if the input exceeds the token budget, we remove whole turns starting from the oldest to meet the budget requirement. \textbf{The answer record is stored independently and remains available after observation history is pruned.}

We can also archive historical frames at lower resolution while keeping recent frames at high resolution. Our experiments show that this reduces visual tokens while maintaining comparable performance.

\section{Experiments}
\label{sec:experiments}

Our experiments compare \method{} with existing proactive video understanding methods on three benchmarks covering response timing and answer quality. We also examine performance across backbone sizes and evaluate inference efficiency under the same hardware conditions.

\subsection{Experimental Setup}
\label{sec:exp-setup}

\paragraph{Benchmarks.}
We evaluate \method{} on three proactive video benchmarks: \textit{StreamingBench}~\cite{lin2026streamingbench} (Proactive Output track, measured by first-response Acc@2s with a $\pm 2\text{s}$ tolerance), \textit{ProactiveVideoQA}~\cite{wang2025proactivevideoqa} (measured by $\mathrm{PAUC}$ with $\omega=0.5$), and \textit{OVO-Bench}~\cite{niu2025ovo} (Forward Active Responding track, covering CRR, SSR, and REC, measured by online F1 with a $\pm 2\text{s}$ tolerance~\cite{zheng2026garde,tian2026syncpro}).

\paragraph{Evaluation Metrics.}
StreamingBench-PO counts a hit when the first answer is within two seconds of the reference time.
For $N_{\mathrm{hit}}$ hits among $N_{\mathrm{total}}$ queries,
\begin{equation}
\mathrm{Acc@2s}=\frac{N_{\mathrm{hit}}}{N_{\mathrm{total}}}.
\label{eq:po-accuracy}
\end{equation}
On OVO-Bench, responses and reference events are matched one-to-one within two seconds.
For each subtask, $P$ and $R$ are precision and recall averaged over queries.
The subtask score is
\begin{equation}
F_1=\frac{2PR}{P+R}.
\label{eq:ovo-f1}
\end{equation}
The final score averages F1 across CRR, SSR, and REC.

On ProactiveVideoQA, GPT-4.1~\cite{gpt41} scores accumulated answers over time.
PAUC is the normalized area under the score curve over time ($\omega=0.5$)~\cite{wang2025proactivevideoqa}.
We equally average the four domain scores.

\paragraph{Implementation Details.}
We use a frozen Qwen3-VL-4B-Instruct~\cite{bai2025qwen3} as our default backbone and evaluate the 2B, 4B, 8B and 32B variants to study model scaling. Video streams are sampled at 1 FPS. At each second, \method{} feeds the VLM the latest $N=4$ high-resolution frames, historical observations archived every 2 seconds, and the answer record. All models use greedy decoding. The full prompts and additional implementation details will be provided in the supplementary material.

\begin{table*}[t]
\centering
\caption{Results on OVO-Bench Forward Active Responding. Precision (P), recall (R), and F1 use a 2\,s tolerance across CRR, SSR, and REC. All scores are higher-is-better percentages; best results are \textbf{bold}; second best are \underline{underlined}.}
\label{tab:ovo_results}
\vspace{2pt}
\footnotesize
\setlength{\tabcolsep}{3pt}
\renewcommand{\arraystretch}{1.05}
\papertablestyle
\begin{tabularx}{\textwidth}{@{}l c|
  *{3}{>{\centering\arraybackslash}X}|
  *{3}{>{\centering\arraybackslash}X}|
  *{3}{>{\centering\arraybackslash}X}|
  >{\centering\arraybackslash}X@{}}
\toprule
\multirow{2}{*}{\textbf{Method}} & \multirow{2}{*}{\textbf{FPS}} & \multicolumn{3}{c|}{\textbf{CRR}} & \multicolumn{3}{c|}{\textbf{SSR}} & \multicolumn{3}{c|}{\textbf{REC}} & \multirow{2}{*}{\shortstack{\textbf{Avg.}\\\textbf{F1}}} \\
\cmidrule(lr){3-5} \cmidrule(lr){6-8} \cmidrule(lr){9-11}
& & P & R & F1 & P & R & F1 & P & R & F1 & \\
\midrule
\multicolumn{12}{@{}c@{}}{\color{black!55}Streaming models} \\[1pt]
VideoLLM-online~\cite{chen2024videollm} & 2 & 2.08 & 2.08 & 2.08 & 6.88 & 11.07 & 8.49 & 16.46 & 7.43 & 10.24 & 6.93 \\
Flash-VStream~\cite{zhang2024flash} & 2 & 5.21 & 6.25 & 5.68 & 9.52 & 2.46 & 3.91 & 7.49 & 3.46 & 4.73 & 4.77 \\
StreamForest~\cite{zeng2026streamforest} & 2 & 2.08 & 2.08 & 2.08 & 13.69 & 8.02 & 10.11 & 28.90 & 30.49 & 29.67 & 13.95 \\
MMDuet2~\cite{wang2026mmduet2} & 2 & 8.47 & 47.92 & 14.40 & -- & -- & -- & \underline{42.72} & 19.33 & 26.62 & -- \\
Em-Garde~\cite{zheng2026garde} & 2 & \textbf{18.22} & 47.92 & 26.40 & 14.87 & \underline{54.88} & 23.40 & 39.66 & 47.33 & 43.16 & 30.99 \\
\midrule
\multicolumn{12}{@{}c@{}}{\color{black!55}Reasoning-enabled models} \\[1pt]
VST~\cite{guan2026video} & 2 & 4.65 & 18.75 & 7.46 & 3.80 & 2.86 & 3.26 & 9.63 & \textbf{95.00} & 17.48 & 9.40 \\
ThinkStream~\cite{liu2026thinking} & 2 & 3.09 & 18.75 & 5.30 & 3.65 & 49.60 & 6.80 & 31.48 & 23.01 & 26.58 & 12.89 \\
SyncPro-SFT~\cite{tian2026syncpro} & 1 & \underline{18.09} & 62.50 & \textbf{28.06} & 13.14 & \textbf{57.06} & 21.36 & 33.12 & \underline{85.14} & 47.69 & 32.37 \\
SyncPro-RL~\cite{tian2026syncpro} & 1 & 16.99 & \underline{64.58} & \underline{26.91} & \textbf{17.50} & 48.17 & \textbf{25.67} & 35.43 & 84.95 & \underline{50.00} & \underline{34.19} \\
\midrule
\rowcolor{tablerowgreen}
\textbf{\method{}} (Ours) & 1 & 16.45 & \textbf{66.67} & 26.39 & \underline{16.19} & 49.88 & \underline{24.44} & \textbf{43.49} & 74.30 & \textbf{54.86} & \textbf{35.23} \\
\bottomrule
\end{tabularx}
\end{table*}

\subsection{Main Results}
\label{sec:main-results}

We compare \method{} with streaming and reasoning-enabled models. Representative approaches learn response timing through multi-turn reinforcement learning (MMDuet2~\cite{wang2026mmduet2}), trigger responses through visual proposal matching (Em-Garde~\cite{zheng2026garde}), or train separate reasoning and response branches (SyncPro~\cite{tian2026syncpro}).

\paragraph{StreamingBench and ProactiveVideoQA.}
As shown in Table~\ref{tab:main_results}, \method{} achieves \textbf{54.80\%} Acc@2s on StreamingBench-PO, exceeding SyncPro-RL by \textbf{13.60} percentage points.
On ProactiveVideoQA, it achieves a mean PAUC of \textbf{52.18}, surpassing MMDuet2 by \textbf{12.38} points, with the best scores on three of the four subsets.
These results demonstrate strong performance in both response timing and answer quality over time, without task-specific training or auxiliary triggering modules.

\paragraph{OVO-Bench Forward Active Responding.}
Table~\ref{tab:ovo_results} shows that \method{} achieves \textbf{35.23} average F1 with a frozen 4B backbone, outperforming Em-Garde and SyncPro-RL.
The gain is driven by REC, where \method{} achieves an F1 of 54.86 compared with 50.00 for SyncPro-RL. It remains competitive on CRR and SSR.
This extends the results to multi-event settings, where the system must track successive or recurring events rather than a single response opportunity.

\subsection{Model Scaling and Inference Efficiency}
\label{sec:scaling-efficiency}

\begin{figure}[!htb]
\centering
\includegraphics[width=\linewidth]{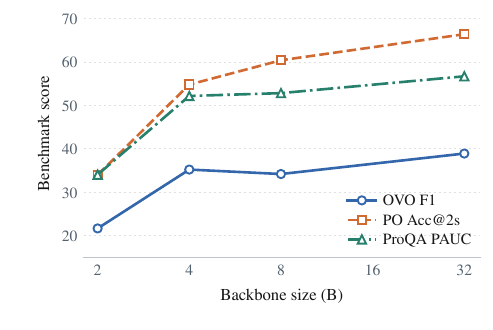}
\caption{Scaling frozen backbones. Scores versus backbone size on a logarithmic axis.
}
\label{fig:backbone-scaling}
\end{figure}

\paragraph{Scaling across Model Sizes.}
Figure~\ref{fig:backbone-scaling} evaluates frozen Qwen3-VL backbones from 2B to 32B, with all variants sharing the same 4B mode initialization.
StreamingBench-PO accuracy increases from 34.0\% to 66.4\%, while ProactiveVideoQA PAUC rises from 34.05 to 56.71.
OVO-Bench shows a less consistent trend: F1 increases from 21.67 at 2B to 35.23 at 4B, drops to 34.20 at 8B, and reaches its highest value of \textbf{38.92} at 32B.
Overall, larger backbones improve performance without task-specific training, although the gains vary across benchmarks.

\begin{samepage}
\paragraph{Runtime Efficiency.}
We compare \method{} (4B) with MMDuet2 (3B) on a single H100 80GB GPU with TP=1 and a single stream for both methods.
\method{} takes \textbf{526.6\,ms} per decision on average, compared with 557.1\,ms for MMDuet2, maintaining comparable decision latency despite using a larger backbone.
\par
\end{samepage}

\section{Analysis}
\label{sec:analysis}

With a frozen VLM, proactive responses depend on the evidence available to the model and the instructions that guide its decisions. We first examine how reporting rules shape response behavior. We then study what memory to retain and how to manage visual history. Finally, we evaluate whether filtering duplicate answers improves the delivered responses.

\subsection{What Makes an Effective Proactive Prompt?}
\label{sec:analysis-prompt}

Modern VLMs perform well on offline long-video understanding benchmarks~\cite{bai2025qwen3,fu2025video,wu2024longvideobench}, but understanding video content does not by itself specify when to answer.
In proactive streaming, the model must decide whether the evidence is sufficient, whether an event has already been reported, and whether a new occurrence calls for another response.

The MMDuet2 prompt~\cite{wang2026mmduet2} asks the model to answer from recent frames or output ``NO REPLY'' when it cannot answer.
This instruction addresses video segments with insufficient evidence. It does not explicitly specify the reporting requirements for different question types or distinguish a new event from the continuation of a reported event.
\par
\begin{samepage}
\noindent Our prompt adds three rules:
\begin{enumerate}[label=\arabic*., leftmargin=1.4em, labelindent=0pt, labelwidth=1em, labelsep=0.4em, align=left, itemsep=3pt, topsep=3pt, parsep=0pt, partopsep=0pt]
    \item \textbf{Specify reporting requirements.} Single-target questions call for an answer when evidence is sufficient; multi-step or recurring-event questions call for a report of each distinct step or occurrence.
    \item \textbf{Avoid repeated reports.} Preparatory actions are not sufficient to trigger an answer, and the continuation of a reported event should not trigger another response.
    \item \textbf{Allow corrections.} Earlier answers are not treated as verified facts. The model can revise them based on additional evidence observed later.
\end{enumerate}
\end{samepage}
We use the original MMDuet2 prompt as a training-free comparison baseline, keeping the 4B backbone and streaming pipeline unchanged (Table~\ref{tab:ablation-prompt}).
With the original MMDuet2 prompt, OVO-Bench F1 drops from \textbf{35.23} to 16.41, and StreamingBench-PO accuracy falls from \textbf{54.8\%} to 42.0\%.
The largest drops occur on SSR and REC, where F1 decreases from 24.44 and 54.86 to 4.12 and 20.00, respectively.
These results show that explicit reporting requirements help the model decide when to answer and when to remain silent.

\begin{table}[!htb]
  \centering
  \caption{Prompt comparison on full OVO-Bench and StreamingBench-PO with the same 4B backbone, 1-second decision interval, and delivery filter.}
  \label{tab:ablation-prompt}
  \vspace{2pt}
  \small
  \setlength{\tabcolsep}{2pt}
  \renewcommand{\arraystretch}{1.20}
  \papertablestyle
  \begin{tabularx}{\columnwidth}{@{}>{\raggedright\arraybackslash}X|cccc|c@{}}
    \toprule
    \textbf{Variant} & \textbf{CRR} & \textbf{SSR} & \textbf{REC} & \textbf{OVO F1} & \textbf{PO Acc.} \\
    \midrule
    MMDuet2 prompt & 25.11 & 4.12 & 20.00 & 16.41 & 42.0 \\
    \rowcolor{tablerowgreen}
    \textbf{\method{} prompt} & \textbf{26.39} & \textbf{24.44} & \textbf{54.86} & \textbf{35.23} & \textbf{54.8} \\
    \bottomrule
  \end{tabularx}
\end{table}

The larger gains on SSR and REC are consistent with their need for continued reporting.
These simple reporting rules also improve first-response accuracy on PO.

\subsection{How Should Memory Be Organized?}
\label{sec:analysis-memory}

Streaming interactions leave different kinds of information: visual observations, \texttt{no} decisions, and answers. Their roles in subsequent responses may differ. We first measure the benefit of retaining a separate answer record, then compare text and image history with this record retained. We examine history length and image resolution separately in Section~\ref{sec:analysis-history-budget}.

\begin{table}[!htb]
  \centering
  \caption{Memory configurations. Tokens are average input length per decision. LR and HR denote low- and high-resolution history.
  }
  \label{tab:ablation-context}
  \vspace{2pt}
  \small
  \setlength{\tabcolsep}{2pt}
  \renewcommand{\arraystretch}{1.20}
  \papertablestyle
  \begin{tabularx}{\columnwidth}{@{}>{\raggedright\arraybackslash}X|c|cc@{}}
    \toprule
    \textbf{Memory configuration} & \textbf{Tokens} & \textbf{OVO F1} & \textbf{PO Acc.} \\
    \midrule
    No memory & \textbf{1.1K} & 18.94 & 59.2 \\
    Images only & 6.3K & 25.55 & 55.2 \\
    Answer record only & 1.8K & 26.80 & \textbf{59.6} \\
    Answer record + text & 3.8K & 29.15 & 52.0 \\
    Answer record + LR images & 5.3K & \textbf{35.29} & 54.4 \\
    \rowcolor{tablerowgreen}
    \textbf{Answer record + HR images} & 8.3K & 35.23 & 54.8 \\
    \bottomrule
  \end{tabularx}
\end{table}

\begin{figure}[!t]
\centering
\captionsetup[subfigure]{position=top,justification=raggedright,singlelinecheck=false,font=small,labelfont=bf,skip=2pt}
\begin{subfigure}{\columnwidth}
\caption{Image memory.}
\includegraphics[page=1,width=\linewidth]{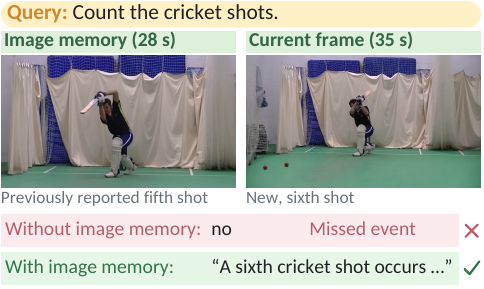}
\end{subfigure}
\par\vspace{5pt}
\begin{subfigure}{\columnwidth}
\caption{Answer record.}
\includegraphics[page=2,width=\linewidth]{figures/cases/memory_behavior_cases.pdf}
\end{subfigure}
\caption{Illustrative cases of complementary memory roles.
(a) Image memory distinguishes a new cricket shot from a previously reported shot.
(b) The answer record preserves an earlier cap-opening report and avoids reporting the same event again.}
\label{fig:memory-cases}
\end{figure}

\paragraph{Role of the Answer Record.}
The model needs to know what it has already answered as well as what the current frames show.
As shown in Table~\ref{tab:ablation-context}, adding only the answer record $A_t$ raises OVO-Bench F1 from 18.94 without memory to \textbf{26.80}.
Even with image history retained, removing $A_t$ lowers F1 from 35.23 to 25.55.
\textbf{Historical images do not replace prior answers, which provide a separate source of information for subsequent response decisions.}
For example, later frames in the video still show the marker cap after it has been opened (\mbox{Figure~\ref{fig:memory-cases}(b)}).
The answer record preserves the earlier opening report and its timestamp after observation pruning.
This gives the model a reference for deciding whether the current scene presents another reporting opportunity.
In this example, the same opening event calls for silence.

\paragraph{Comparing Text and Image Memory.}
Remembering prior answers does not preserve the scenes that preceded them.
With $A_t$ retained, adding text history raises F1 from 26.80 to \textbf{29.15}, while retaining historical images raises it further to \textbf{35.23} (Table~\ref{tab:ablation-context}).
The many \texttt{no} replies in past interactions record that the model did not answer, but do not describe what was visible at the time.
Image history supplies this information and lets the model compare observations before and after an event.
The two forms of memory preserve what the model has said and what it has seen, respectively, supporting subsequent decisions together.
For example, the current frame shows a cricket shot (\mbox{Figure~\ref{fig:memory-cases}(a)}).
The earlier frame records the previously reported fifth shot, allowing the model to identify the current action as a new sixth shot.
Historical context is less helpful on PO, which evaluates first-response timing rather than continued reporting. Recent frames can often provide sufficient evidence for a timely first response.

\subsection{How Should Visual History Be Managed?}
\label{sec:analysis-history-budget}

Interaction history grows as video arrives. Under a fixed 32K context limit, retaining more history increases input cost. We vary the number of retained turns and the resolution of historical images to examine how to allocate this limited context budget.

\paragraph{History Retention and Pruning.}
We vary the turn-count threshold for pruning and the number of recent turns retained afterward, following the rule defined in Section~\ref{sec:memory-management}.
For example, $12{\to}8$ means retaining the latest eight turns when history reaches twelve turns. The current interaction is then appended.
Table~\ref{tab:context-efficiency} compares the input cost and performance of these configurations.
Untrimmed history and longer retention windows use 9.5K–12.1K input tokens per step on average and yield OVO-Bench F1 scores of 23.53–34.01.
The default 12→8 policy uses 8.3K tokens on average and achieves 35.23 F1.

\begin{table}[!htb]
  \centering
  \caption{History cost and performance under a 32K context limit. Tokens are average input length per decision. HR and LR denote high- and low-resolution history.}
  \label{tab:context-efficiency}
  \vspace{2pt}
  \small
  \setlength{\tabcolsep}{3pt}
  \renewcommand{\arraystretch}{1.20}
  \papertablestyle
  \begin{tabularx}{\columnwidth}{@{}>{\raggedright\arraybackslash}X|c|cc@{}}
    \toprule
    \textbf{History policy} & \textbf{Tokens} & \textbf{OVO F1} & \textbf{PO Acc.} \\
    \midrule
    No pruning & 10.8K & 23.53 & 54.0 \\
    $24{\to}16$ & 12.1K & 32.88 & \textbf{54.8} \\
    $16{\to}8$ & 9.5K & 34.01 & \textbf{54.8} \\
    \rowcolor{tablerowgreen}
    \textbf{$12{\to}8$, HR (default)} & 8.3K & 35.23 & \textbf{54.8} \\
    $12{\to}8$, LR & \textbf{5.3K} & \textbf{35.29} & 54.4 \\
    \bottomrule
  \end{tabularx}
\end{table}

Retaining longer history does not improve F1 in these comparisons and requires more input tokens.
Meanwhile, the separate answer record preserves prior answers and their timestamps, so pruning observations does not remove these records.
Within this budget, limiting observation history while retaining the answer record performs better than accumulating the full interaction history.

\paragraph{Historical Image Resolution.}
With the default retention policy, we keep recent frames at high resolution and downsample only historical images.
Average input length decreases from 8.3K to \textbf{5.3K tokens}, a reduction of \textbf{36.3\%}, while F1 changes from 35.23 to 35.29.
The nearly unchanged F1 indicates that lower-resolution images still provide useful historical references.
Recent frames retain their original visual detail.
StreamingBench-PO accuracy remains similar, changing from 54.8\% to 54.4\%.

\subsection{Does Duplicate Filtering Help?}
\label{sec:analysis-cadence}

We find that, in multi-turn conversations, the model still tends to generate consecutive identical answers, even with instructions to avoid repeated reports and access to prior answer records. We therefore filter these answers before delivery while leaving the model's internal history unchanged.

\begin{table}[!htb]
  \centering
  \caption{Duplicate-response filtering on full OVO-Bench and StreamingBench-PO at a 1-second decision interval.}
  \label{tab:ablation-cadence}
  \vspace{2pt}
  \small
  \setlength{\tabcolsep}{2pt}
  \renewcommand{\arraystretch}{1.20}
  \papertablestyle
  \begin{tabularx}{\columnwidth}{@{}>{\raggedright\arraybackslash}X|cccc|c@{}}
    \toprule
    \textbf{Variant} & \textbf{CRR} & \textbf{SSR} & \textbf{REC} & \textbf{OVO F1} & \textbf{PO Acc.} \\
    \midrule
    Without filtering & 23.17 & 19.61 & 46.64 & 29.81 & \textbf{54.8} \\
    \rowcolor{tablerowgreen}
    \textbf{With filtering} & \textbf{26.39} & \textbf{24.44} & \textbf{54.86} & \textbf{35.23} & \textbf{54.8} \\
    \bottomrule
  \end{tabularx}
\end{table}

As shown in Table~\ref{tab:ablation-cadence}, duplicate filtering raises average OVO-Bench F1 from 29.81 to \textbf{35.23}. CRR, SSR, and REC increase from 23.17, 19.61, and 46.64 to 26.39, 24.44, and 54.86, respectively. Removing redundant answers consistently improves performance across all three subtasks. StreamingBench-PO accuracy remains unchanged at 54.8\%.

The answer record guides subsequent response decisions, while the filter removes consecutive identical answers that the model still generates. They reduce repeated deliveries together, while retaining raw answers for future decisions.

\section{Conclusion}
\label{sec:conclusion}

We present \method{}, a training-free baseline for proactive video understanding that adapts an off-the-shelf VLM to continuous streaming interaction by organizing recent high-resolution frames, bounded historical observations, and a separately retained answer record.
With single-step joint decision generation and duplicate filtering, \method{} surpasses specialized RL-trained and trigger-based proactive systems across OVO-Bench, StreamingBench-PO, and ProactiveVideoQA, and shows performance gains as backbone size increases from 2B to 32B without retraining.
Our analysis highlights the complementary roles of visual history and prior answers. Compressing historical images reduces input cost while maintaining comparable performance.

\label{main-text-end}
\section*{LLM Usage Statement}
ChatGPT assisted with manuscript drafting and editing, literature organization, formulation of diagnostic analyses, and preparation of the schematic and document source. It was not used to generate experimental results. Model-based evaluation of ProactiveVideoQA is described separately in the evaluation protocol.

{\small
\begin{thebibliography}{61}
\providecommand{\natexlab}[1]{#1}
\providecommand{\url}[1]{\texttt{#1}}
\expandafter\ifx\csname urlstyle\endcsname\relax
  \providecommand{\doi}[1]{doi: #1}\else
  \providecommand{\doi}{doi: \begingroup \urlstyle{rm}\Url}\fi

\bibitem[Alayrac et~al.(2022)Alayrac, Donahue, Luc, Miech, Barr, Hasson, Lenc,
  Mensch, Millican, Reynolds, et~al.]{alayrac2022flamingo}
Jean-Baptiste Alayrac, Jeff Donahue, Pauline Luc, Antoine Miech, Iain Barr,
  Yana Hasson, Karel Lenc, Arthur Mensch, Katherine Millican, Malcolm Reynolds,
  et~al.
\newblock {Flamingo}: a visual language model for few-shot learning.
\newblock \emph{Advances in neural information processing systems},
  35:\penalty0 23716--23736, 2022.

\bibitem[Azad et~al.(2026)Azad, Vineet, and Rawat]{azad2026streamready}
Shehreen Azad, Vibhav Vineet, and Yogesh~S Rawat.
\newblock {StreamReady}: Learning what to answer and when in long streaming
  videos.
\newblock In \emph{Proceedings of the IEEE/CVF Conference on Computer Vision
  and Pattern Recognition}, pages 40494--40504, 2026.

\bibitem[Bai et~al.(2025{\natexlab{a}})Bai, Cai, Chen, Chen, Chen, Cheng, Deng,
  Ding, Gao, Ge, et~al.]{bai2025qwen3}
Shuai Bai, Yuxuan Cai, Ruizhe Chen, Keqin Chen, Xionghui Chen, Zesen Cheng,
  Lianghao Deng, Wei Ding, Chang Gao, Chunjiang Ge, et~al.
\newblock Qwen3-vl technical report.
\newblock \emph{arXiv preprint arXiv:2511.21631}, 2025{\natexlab{a}}.

\bibitem[Bai et~al.(2025{\natexlab{b}})Bai, Chen, Liu, Wang, Ge, Song, Dang,
  Wang, Wang, Tang, Zhong, Zhu, Yang, Li, Wan, Wang, Ding, Fu, Xu, Ye, Zhang,
  Xie, Cheng, Zhang, Yang, Xu, and Lin]{bai2025qwen25vltechnicalreport}
Shuai Bai, Keqin Chen, Xuejing Liu, Jialin Wang, Wenbin Ge, Sibo Song, Kai
  Dang, Peng Wang, Shijie Wang, Jun Tang, Humen Zhong, Yuanzhi Zhu, Mingkun
  Yang, Zhaohai Li, Jianqiang Wan, Pengfei Wang, Wei Ding, Zheren Fu, Yiheng
  Xu, Jiabo Ye, Xi Zhang, Tianbao Xie, Zesen Cheng, Hang Zhang, Zhibo Yang,
  Haiyang Xu, and Junyang Lin.
\newblock Qwen2.5-vl technical report, 2025{\natexlab{b}}.

\bibitem[Chen et~al.(2024)Chen, Lv, Wu, Lin, Song, Gao, Liu, Gao, Mao, and
  Shou]{chen2024videollm}
Joya Chen, Zhaoyang Lv, Shiwei Wu, Kevin~Qinghong Lin, Chenan Song, Difei Gao,
  Jia-Wei Liu, Ziteng Gao, Dongxing Mao, and Mike~Zheng Shou.
\newblock {VideoLLM-online}: Online video large language model for streaming
  video.
\newblock In \emph{2024 IEEE/CVF Conference on Computer Vision and Pattern
  Recognition (CVPR)}, pages 18407--18418. IEEE, 2024.

\bibitem[Di et~al.(2025)Di, Yu, Zhang, Li, Cheng, Li, He, Shu, and
  Jiang]{di2025streaming}
Shangzhe Di, Zhelun Yu, Guanghao Zhang, Haoyuan Li, Hao Cheng, Bolin Li,
  Wanggui He, Fangxun Shu, and Hao Jiang.
\newblock Streaming video question-answering with in-context video {KV-Cache}
  retrieval.
\newblock In \emph{International Conference on Learning Representations}, pages
  42115--42127, 2025.

\bibitem[Ding et~al.(2025)Ding, Wu, Yang, Jiang, Zhang, Bai, Chen, and
  Cao]{ding2025streammind}
Xin Ding, Hao Wu, Yifan Yang, Shiqi Jiang, Qianxi Zhang, Donglin Bai, Zhibo
  Chen, and Ting Cao.
\newblock {StreamMind}: Unlocking full frame rate streaming video dialogue
  through event-gated cognition.
\newblock In \emph{2025 IEEE/CVF International Conference on Computer Vision
  (ICCV)}, pages 13448--13459. IEEE, 2025.

\bibitem[Fu et~al.(2025)Fu, Dai, Luo, Li, Ren, Zhang, Wang, Zhou, Shen, Zhang,
  et~al.]{fu2025video}
Chaoyou Fu, Yuhan Dai, Yongdong Luo, Lei Li, Shuhuai Ren, Renrui Zhang, Zihan
  Wang, Chenyu Zhou, Yunhang Shen, Mengdan Zhang, et~al.
\newblock {Video-MME}: The first-ever comprehensive evaluation benchmark of
  multi-modal {LLMs} in video analysis.
\newblock In \emph{2025 IEEE/CVF Conference on Computer Vision and Pattern
  Recognition (CVPR)}, pages 24108--24118. IEEE, 2025.

\bibitem[Guan et~al.(2026)Guan, Yin, Liang, Ju, Luo, Luan, Liu, and
  Bai]{guan2026video}
Yiran Guan, Liang Yin, Dingkang Liang, Jianzhong Ju, Zhenbo Luo, Jian Luan,
  Yuliang Liu, and Xiang Bai.
\newblock Video streaming thinking: {VideoLLMs} can watch and think
  simultaneously.
\newblock In \emph{European Conference on Computer Vision}, pages 80--99.
  Springer, 2026.

\bibitem[Huang et~al.(2025)Huang, Li, Li, Wang, Zeng, Liang, Wu, Chen, Li, and
  Wang]{huang2025online}
Zhenpeng Huang, Xinhao Li, Jiaqi Li, Jing Wang, Xiangyu Zeng, Cheng Liang, Tao
  Wu, Xi Chen, Liang Li, and Limin Wang.
\newblock Online video understanding: {OVBench} and {VideoChat-Online}.
\newblock In \emph{2025 IEEE/CVF Conference on Computer Vision and Pattern
  Recognition (CVPR)}, pages 3328--3338. IEEE, 2025.

\bibitem[Jin et~al.(2025)Jin, Yu, Yu, Liu, Liu, Tao, Pei, Wang, Dang, Liu,
  et~al.]{jin2025streamingassistant}
Xinqi Jin, Hanxun Yu, Bohan Yu, Kebin Liu, Jian Liu, Keda Tao, Yixuan Pei, Huan
  Wang, Fan Dang, Jiangchuan Liu, et~al.
\newblock {StreamingAssistant}: Efficient visual token pruning for accelerating
  online video understanding.
\newblock \emph{arXiv preprint arXiv:2512.12560}, 2025.

\bibitem[Li et~al.(2026{\natexlab{a}})Li, Xiao, Yue, Xu, Yao, Li, Fu, Ju, Luan,
  and Jin]{li2026streampro}
Ao Li, Zihan Xiao, Zihao Yue, Boshen Xu, Linli Yao, Jiaze Li, Pei Fu, Jianzhong
  Ju, Jian Luan, and Qin Jin.
\newblock {StreamPro}: From reactive perception to proactive decision-making in
  streaming video.
\newblock \emph{arXiv preprint arXiv:2605.16381}, 2026{\natexlab{a}}.

\bibitem[Li et~al.(2024{\natexlab{a}})Li, Zhang, Guo, Zhang, Li, Zhang, Zhang,
  Zhang, Li, Liu, et~al.]{li2024llava}
Bo Li, Yuanhan Zhang, Dong Guo, Renrui Zhang, Feng Li, Hao Zhang, Kaichen
  Zhang, Peiyuan Zhang, Yanwei Li, Ziwei Liu, et~al.
\newblock {LLaVA-OneVision}: Easy visual task transfer.
\newblock \emph{arXiv preprint arXiv:2408.03326}, 2024{\natexlab{a}}.

\bibitem[Li et~al.(2023)Li, Li, Savarese, and Hoi]{li2023blip}
Junnan Li, Dongxu Li, Silvio Savarese, and Steven Hoi.
\newblock {BLIP-2}: Bootstrapping language-image pre-training with frozen image
  encoders and large language models.
\newblock In \emph{International conference on machine learning}, pages
  19730--19742. PmLR, 2023.

\bibitem[Li et~al.(2024{\natexlab{b}})Li, Wang, He, Li, Wang, Liu, Wang, Xu,
  Chen, Luo, et~al.]{li2024mvbench}
Kunchang Li, Yali Wang, Yinan He, Yizhuo Li, Yi Wang, Yi Liu, Zun Wang, Jilan
  Xu, Guo Chen, Ping Luo, et~al.
\newblock {MVBench}: A comprehensive multi-modal video understanding benchmark.
\newblock In \emph{Proceedings of the IEEE/CVF conference on computer vision
  and pattern recognition}, pages 22195--22206, 2024{\natexlab{b}}.

\bibitem[Li et~al.(2025)Li, He, Wang, Li, Wang, Luo, Wang, Wang, and
  Qiao]{li2025videochat}
KunChang Li, Yinan He, Yi Wang, Yizhuo Li, Wenhai Wang, Ping Luo, Yali Wang,
  Limin Wang, and Yu Qiao.
\newblock {VideoChat}: Chat-centric video understanding.
\newblock \emph{Science China Information Sciences}, 68\penalty0 (10):\penalty0
  200102, 2025.

\bibitem[Li et~al.(2026{\natexlab{b}})Li, Zhu, Zeng, Dong, Wu, Zhang, Yang, Ma,
  Zhang, Shi, et~al.]{li2026videochat3}
Xinhao Li, Yuhan Zhu, Xiangyu Zeng, Yuhao Dong, Haoning Wu, Zhiqiu Zhang,
  Yuandong Yang, Changlian Ma, Qingyu Zhang, Yansong Shi, et~al.
\newblock {VideoChat3}: Fully open video {MLLM} for efficient and generalist
  video understanding.
\newblock \emph{arXiv preprint arXiv:2607.14935}, 2026{\natexlab{b}}.

\bibitem[Li et~al.(2024{\natexlab{c}})Li, Wang, and Jia]{li2024llama}
Yanwei Li, Chengyao Wang, and Jiaya Jia.
\newblock {LLaMA-VID}: An image is worth 2 tokens in large language models.
\newblock In \emph{European conference on computer vision}, pages 323--340.
  Springer, 2024{\natexlab{c}}.

\bibitem[Lin et~al.(2024)Lin, Ye, Zhu, Cui, Ning, Jin, and Yuan]{lin2024video}
Bin Lin, Yang Ye, Bin Zhu, Jiaxi Cui, Munan Ning, Peng Jin, and Li Yuan.
\newblock {Video-LLaVA}: Learning united visual representation by alignment
  before projection.
\newblock In \emph{Proceedings of the 2024 conference on empirical methods in
  natural language processing}, pages 5971--5984, 2024.

\bibitem[Lin et~al.(2026)Lin, Fang, Chen, Cheng, Wan, Luo, Wang, Li, Liu, and
  Sun]{lin2026streamingbench}
Junming Lin, Zheng Fang, Chi Chen, Haoxuan Cheng, Zihao Wan, Fuwen Luo, Ziyue
  Wang, Peng Li, Yang Liu, and Maosong Sun.
\newblock Streamingbench: Assessing the gap for mllms to achieve streaming
  video understanding.
\newblock In \emph{ICASSP 2026-2026 IEEE International Conference on Acoustics,
  Speech and Signal Processing (ICASSP)}, pages 12147--12151. IEEE, 2026.

\bibitem[Liu et~al.(2023)Liu, Li, Wu, and Lee]{liu2023visual}
Haotian Liu, Chunyuan Li, Qingyang Wu, and Yong~Jae Lee.
\newblock Visual instruction tuning.
\newblock \emph{Advances in neural information processing systems},
  36:\penalty0 34892--34916, 2023.

\bibitem[Liu et~al.(2025)Liu, Zhao, Xu, and Ghanem]{liu2025bolt}
Shuming Liu, Chen Zhao, Tianqi Xu, and Bernard Ghanem.
\newblock {BOLT}: Boost large vision-language model without training for
  long-form video understanding.
\newblock In \emph{2025 IEEE/CVF Conference on Computer Vision and Pattern
  Recognition (CVPR)}, pages 3318--3327. IEEE, 2025.

\bibitem[Liu et~al.(2026)Liu, Guo, Li, Zhen, He, Ji, Ren, Zhang, Lu, and
  Liu]{liu2026thinking}
Zikang Liu, Longteng Guo, Handong Li, Ru Zhen, Xingjian He, Ruyi Ji, Xiaoming
  Ren, Yanhao Zhang, Haonan Lu, and Jing Liu.
\newblock Thinking in streaming video.
\newblock In \emph{European Conference on Computer Vision}, pages 21--38.
  Springer, 2026.

\bibitem[Lu et~al.(2026)Lu, Guan, Bo, Chen, Guo, Li, Liu, Sun, Li, Zhang,
  et~al.]{lu2026phostream}
Xudong Lu, Huankang Guan, Yang Bo, Jinpeng Chen, Xintong Guo, Shuhan Li, Fang
  Liu, Peiwen Sun, Xueying Li, Wei Zhang, et~al.
\newblock {PhoStream}: Benchmarking real-world streaming for omnimodal
  assistants in mobile scenarios.
\newblock \emph{arXiv preprint arXiv:2601.22575}, 2026.

\bibitem[Maaz et~al.(2024)Maaz, Rasheed, Khan, and Khan]{maaz2024video}
Muhammad Maaz, Hanoona Rasheed, Salman Khan, and Fahad Khan.
\newblock {Video-ChatGPT}: Towards detailed video understanding via large
  vision and language models.
\newblock In \emph{Proceedings of the 62nd Annual Meeting of the Association
  for Computational Linguistics (Volume 1: Long Papers)}, pages 12585--12602,
  2024.

\bibitem[Niu et~al.(2025)Niu, Li, Miao, Ge, Zhou, He, Dong, Duan, Ding, Qian,
  et~al.]{niu2025ovo}
Junbo Niu, Yifei Li, Ziyang Miao, Chunjiang Ge, Yuanhang Zhou, Qihao He, Xiaoyi
  Dong, Haodong Duan, Shuangrui Ding, Rui Qian, et~al.
\newblock {OVO-Bench}: How far is your {Video-LLMs} from real-world online
  video understanding?
\newblock In \emph{2025 IEEE/CVF Conference on Computer Vision and Pattern
  Recognition (CVPR)}, pages 18902--18913. IEEE, 2025.

\bibitem[{OpenAI}(2025)]{gpt41}
{OpenAI}.
\newblock Introducing {GPT-4.1} in the {API}.
\newblock \url{https://openai.com/index/gpt-4-1/}, 2025.

\bibitem[Qian et~al.(2024)Qian, Dong, Zhang, Zang, Ding, Lin, and
  Wang]{qian2024streaming}
Rui Qian, Xiaoyi Dong, Pan Zhang, Yuhang Zang, Shuangrui Ding, Dahua Lin, and
  Jiaqi Wang.
\newblock Streaming long video understanding with large language models.
\newblock \emph{Advances in Neural Information Processing Systems},
  37:\penalty0 119336--119360, 2024.

\bibitem[Qian et~al.(2025)Qian, Ding, Dong, Zhang, Zang, Cao, Lin, and
  Wang]{qian2025dispider}
Rui Qian, Shuangrui Ding, Xiaoyi Dong, Pan Zhang, Yuhang Zang, Yuhang Cao,
  Dahua Lin, and Jiaqi Wang.
\newblock {Dispider}: Enabling video {LLMs} with active real-time interaction
  via disentangled perception, decision, and reaction.
\newblock In \emph{2025 IEEE/CVF Conference on Computer Vision and Pattern
  Recognition (CVPR)}, pages 24045--24055. IEEE, 2025.

\bibitem[Shen et~al.(2024)Shen, Xiong, Zhao, Wu, Chen, Zhu, Liu, Xiao,
  Varadarajan, Bordes, et~al.]{shen2024longvu}
Xiaoqian Shen, Yunyang Xiong, Changsheng Zhao, Lemeng Wu, Jun Chen, Chenchen
  Zhu, Zechun Liu, Fanyi Xiao, Balakrishnan Varadarajan, Florian Bordes, et~al.
\newblock {LongVU}: Spatiotemporal adaptive compression for long video-language
  understanding.
\newblock \emph{arXiv preprint arXiv:2410.17434}, 2024.

\bibitem[Shen et~al.(2026)Shen, Tian, Yang, and Liu]{shen2026simple}
Yujiao Shen, Shulin Tian, Jingkang Yang, and Ziwei Liu.
\newblock A simple baseline for streaming video understanding.
\newblock \emph{arXiv preprint arXiv:2604.02317}, 2026.

\bibitem[Shi et~al.(2026)Shi, Zhao, Jiang, Zeng, Wang, and Wang]{shi2026river}
Yansong Shi, Qingsong Zhao, Tianxiang Jiang, Xiangyu Zeng, Yi Wang, and Limin
  Wang.
\newblock {RIVER}: A real-time interaction benchmark for video {LLMs}.
\newblock \emph{arXiv preprint arXiv:2603.03985}, 2026.

\bibitem[Song et~al.(2024)Song, Chai, Wang, Zhang, Zhou, Wu, Chi, Guo, Ye,
  Zhang, et~al.]{song2024moviechat}
Enxin Song, Wenhao Chai, Guanhong Wang, Yucheng Zhang, Haoyang Zhou, Feiyang
  Wu, Haozhe Chi, Xun Guo, Tian Ye, Yanting Zhang, et~al.
\newblock {MovieChat}: From dense token to sparse memory for long video
  understanding.
\newblock In \emph{2024 IEEE/CVF Conference on Computer Vision and Pattern
  Recognition (CVPR)}, pages 18221--18232. IEEE, 2024.

\bibitem[Tang et~al.(2025)Tang, Qiu, Xie, Tian, Jiao, and Ye]{tang2025adaptive}
Xi Tang, Jihao Qiu, Lingxi Xie, Yunjie Tian, Jianbin Jiao, and Qixiang Ye.
\newblock Adaptive keyframe sampling for long video understanding.
\newblock In \emph{2025 IEEE/CVF Conference on Computer Vision and Pattern
  Recognition (CVPR)}, pages 29118--29128. IEEE, 2025.

\bibitem[Tian et~al.(2026)Tian, Xu, Liu, Yu, Shi, Zhao, Huang, Wang, and
  Shen]{tian2026syncpro}
Xueyun Tian, Bingbing Xu, Jiaming Liu, Jinpeng Yu, Yang Shi, Yu Zhao, Ruihua
  Huang, Yuanzhuo Wang, and Huawei Shen.
\newblock {SyncPro}: Synchronous reasoning and responding for proactive
  streaming video understanding.
\newblock \emph{arXiv preprint arXiv:2610.32132}, 2026.

\bibitem[Wang et~al.(2026{\natexlab{a}})Wang, Feng, Lai, Xu, Li, Ge, Dehghan,
  Cao, and Huang]{wang2026streambridge}
Haibo Wang, Bo Feng, Zhengfeng Lai, Mingze Xu, Shiyu Li, Weifeng Ge, Afshin
  Dehghan, Meng Cao, and Ping Huang.
\newblock {StreamBridge}: Turning your offline video large language model into
  a proactive streaming assistant.
\newblock \emph{Advances in Neural Information Processing Systems},
  38:\penalty0 132332--132359, 2026{\natexlab{a}}.

\bibitem[Wang et~al.(2025{\natexlab{a}})Wang, Gao, Gu, Pu, Cui, Wei, Liu, Jing,
  Ye, Shao, et~al.]{wang2025internvl3}
Weiyun Wang, Zhangwei Gao, Lixin Gu, Hengjun Pu, Long Cui, Xingguang Wei,
  Zhaoyang Liu, Linglin Jing, Shenglong Ye, Jie Shao, et~al.
\newblock {InternVL3.5}: Advancing open-source multimodal models in
  versatility, reasoning, and efficiency.
\newblock \emph{arXiv preprint arXiv:2508.18265}, 2025{\natexlab{a}}.

\bibitem[Wang et~al.(2026{\natexlab{b}})Wang, Huang, Wu, Zhao, Xu, Xia, and
  Peng]{wang2026livibench}
Xiaodong Wang, Langling Huang, Zhirong Wu, Xu Zhao, Teng Xu, Xuhong Xia, and
  Peixi Peng.
\newblock {LiViBench}: An omnimodal benchmark for interactive livestream video
  understanding.
\newblock In \emph{Proceedings of the AAAI Conference on Artificial
  Intelligence}, pages 26517--26525, 2026{\natexlab{b}}.

\bibitem[Wang et~al.(2025{\natexlab{b}})Wang, Meng, Wang, Liang, Wei, Zhang,
  and Zhao]{wang2025videollm}
Yueqian Wang, Xiaojun Meng, Yuxuan Wang, Jianxin Liang, Jiansheng Wei, Huishuai
  Zhang, and Dongyan Zhao.
\newblock {VideoLLM} knows when to speak: Enhancing time-sensitive video
  comprehension with video-text duet interaction format.
\newblock In \emph{EMNLP (Findings)}, pages 6338--6359, 2025{\natexlab{b}}.

\bibitem[Wang et~al.(2025{\natexlab{c}})Wang, Meng, Wang, Zhang, and
  Zhao]{wang2025proactivevideoqa}
Yueqian Wang, Xiaojun Meng, Yifan Wang, Huishuai Zhang, and Dongyan Zhao.
\newblock {ProactiveVideoQA}: A comprehensive benchmark evaluating proactive
  interactions in video large language models.
\newblock \emph{arXiv preprint arXiv:2507.09313}, 2025{\natexlab{c}}.

\bibitem[Wang et~al.(2025{\natexlab{d}})Wang, Wang, Chen, Wu, Zhao, and
  Zheng]{wang2025omnimmi}
Yuxuan Wang, Yueqian Wang, Bo Chen, Tong Wu, Dongyan Zhao, and Zilong Zheng.
\newblock {OmniMMI}: A comprehensive multi-modal interaction benchmark in
  streaming video contexts.
\newblock In \emph{2025 IEEE/CVF Conference on Computer Vision and Pattern
  Recognition (CVPR)}, pages 18925--18935. IEEE, 2025{\natexlab{d}}.

\bibitem[Wang et~al.(2026{\natexlab{c}})Wang, Liu, Wang, Xu, Wan, Zhang, and
  Zhao]{wang2026mmduet2}
Yueqian Wang, Songxiang Liu, Disong Wang, Nuo Xu, Guanglu Wan, Huishuai Zhang,
  and Dongyan Zhao.
\newblock {MMDuet2}: Enhancing proactive interaction of video {MLLMs} with
  multi-turn reinforcement learning.
\newblock In \emph{International Conference on Learning Representations}, pages
  23257--23270, 2026{\natexlab{c}}.

\bibitem[Wang et~al.(2026{\natexlab{d}})Wang, Zheng, Mao, Li, Yang, Yu, Yan,
  Yao, and Wang]{wang2026efficient}
Yilin Wang, Xiangxi Zheng, Dongxing Mao, Linjie Li, Zhengyuan Yang, Ping Yu,
  Rui Yan, Yuan Yao, and Alex~Jinpeng Wang.
\newblock Efficient frame selection for long videos at test time with
  attention-based {MLLM} selectors.
\newblock \emph{arXiv preprint arXiv:2607.15689}, 2026{\natexlab{d}}.

\bibitem[Wu et~al.(2024{\natexlab{a}})Wu, Li, Chen, and
  Li]{wu2024longvideobench}
Haoning Wu, Dongxu Li, Bei Chen, and Junnan Li.
\newblock {LongVideoBench}: A benchmark for long-context interleaved
  video-language understanding.
\newblock \emph{Advances in Neural Information Processing Systems},
  37:\penalty0 28828--28857, 2024{\natexlab{a}}.

\bibitem[Wu et~al.(2024{\natexlab{b}})Wu, Chen, Lin, Wang, Gao, Xu, Xu, Hu,
  Chen, and Shou]{wu2024videollm}
Shiwei Wu, Joya Chen, Kevin~Qinghong Lin, Qimeng Wang, Yan Gao, Qianli Xu, Tong
  Xu, Yao Hu, Enhong Chen, and Mike~Zheng Shou.
\newblock {VideoLLM-MoD}: Efficient video-language streaming with
  mixture-of-depths vision computation.
\newblock \emph{Advances in Neural Information Processing Systems},
  37:\penalty0 109922--109947, 2024{\natexlab{b}}.

\bibitem[Xia et~al.(2026)Xia, Chen, Zhang, Sun, and Zhou]{xia2026streaming}
Jiaer Xia, Peixian Chen, Mengdan Zhang, Xing Sun, and Kaiyang Zhou.
\newblock Streaming video instruction tuning.
\newblock In \emph{Proceedings of the IEEE/CVF Conference on Computer Vision
  and Pattern Recognition}, pages 31219--31229, 2026.

\bibitem[Xie et~al.(2026)Xie, He, Wang, Zheng, Ye, and Wu]{xie2026fluxmem}
Yiweng Xie, Bo He, Junke Wang, Xiangyu Zheng, Ziyi Ye, and Zuxuan Wu.
\newblock {FluxMem}: Adaptive hierarchical memory for streaming video
  understanding.
\newblock \emph{arXiv preprint arXiv:2603.02096}, 2026.

\bibitem[Xu et~al.(2026)Xu, Xiao, Chen, He, Peng, Lu, and
  Han]{xu2026streamingvlm}
Ruyi Xu, Guangxuan Xiao, Yukang Chen, Liuning He, Kelly Peng, Yao Lu, and Song
  Han.
\newblock {StreamingVLM}: Real-time understanding for infinite video streams.
\newblock In \emph{International Conference on Learning Representations}, pages
  61463--61475, 2026.

\bibitem[Yang et~al.(2025{\natexlab{a}})Yang, Tang, Zhao, Zhuang, Lu, An, Hu,
  Zhang, Swikir, He, et~al.]{yang2025streamagent}
Haolin Yang, Feilong Tang, Lingxiao Zhao, Xinlin Zhuang, Yifan Lu, Xiang An,
  Ming Hu, Xiaofeng Zhang, Abdalla Swikir, Junjun He, et~al.
\newblock {StreamAgent}: Towards anticipatory agents for streaming video
  understanding.
\newblock \emph{arXiv preprint arXiv:2508.01875}, 2025{\natexlab{a}}.

\bibitem[Yang et~al.(2025{\natexlab{b}})Yang, Zhao, Shukla, Singh, Mishra,
  Zhang, and Ren]{yang2025streammem}
Yanlai Yang, Zhuokai Zhao, Satya~Narayan Shukla, Aashu Singh, Shlok~Kumar
  Mishra, Lizhu Zhang, and Mengye Ren.
\newblock {StreamMem}: Query-agnostic {KV} cache memory for streaming video
  understanding.
\newblock \emph{arXiv preprint arXiv:2508.15717}, 2025{\natexlab{b}}.

\bibitem[Yang et~al.(2025{\natexlab{c}})Yang, Hu, Du, Xue, Qian, Wu, Yang,
  Dong, and Xu]{yang2025svbench}
Zhenyu Yang, Yuhang Hu, Zemin Du, Dizhan Xue, Shengsheng Qian, Jiahong Wu, Fan
  Yang, Weiming Dong, and Changsheng Xu.
\newblock {SVBench}: A benchmark with temporal multi-turn dialogues for
  streaming video understanding.
\newblock In \emph{International Conference on Learning Representations}, pages
  42522--42556, 2025{\natexlab{c}}.

\bibitem[Yang et~al.(2026)Yang, Zhang, Hu, Wang, Qian, Wen, Yang, Gao, Dong,
  and Xu]{yang2026livestar}
Zhenyu Yang, Kairui Zhang, Yuhang Hu, Bing Wang, Shengsheng Qian, Bin Wen, Fan
  Yang, Tingting Gao, Weiming Dong, and Changsheng Xu.
\newblock {LiveStar}: Live streaming assistant for real-world online video
  understanding.
\newblock \emph{Advances in Neural Information Processing Systems},
  38:\penalty0 31266--31304, 2026.

\bibitem[Yao et~al.(2025)Yao, Li, Wei, Li, Ren, Liu, Ouyang, Wang, Li, Li,
  et~al.]{yao2025timechat}
Linli Yao, Yicheng Li, Yuancheng Wei, Lei Li, Shuhuai Ren, Yuanxin Liu, Kun
  Ouyang, Lean Wang, Shicheng Li, Sida Li, et~al.
\newblock {TimeChat-Online}: 80\% visual tokens are naturally redundant in
  streaming videos.
\newblock In \emph{Proceedings of the 33rd ACM International Conference on
  Multimedia}, pages 10807--10816, 2025.

\bibitem[Yu et~al.(2025)Yu, Jin, Wang, Chen, Jin, Zuo, Xu, Sun, Zhang, Wu,
  et~al.]{yu2025frame}
Sicheng Yu, Chengkai Jin, Huanyu Wang, Zhenghao Chen, Sheng Jin, Zhongrong Zuo,
  Xiaolei Xu, Zhenbang Sun, Bingni Zhang, Jiawei Wu, et~al.
\newblock {Frame-Voyager}: Learning to query frames for video large language
  models.
\newblock In \emph{International Conference on Learning Representations}, pages
  84154--84179, 2025.

\bibitem[Zeng et~al.(2026)Zeng, Qiu, Zhang, Li, Wang, Li, Yan, Tian, Tian,
  Zhao, et~al.]{zeng2026streamforest}
Xiangyu Zeng, Kefan Qiu, Qingyu Zhang, Xinhao Li, Jing Wang, Jiaxin Li, Ziang
  Yan, Kun Tian, Meng Tian, Xinhai Zhao, et~al.
\newblock {StreamForest}: Efficient online video understanding with persistent
  event memory.
\newblock \emph{Advances in Neural Information Processing Systems},
  38:\penalty0 75804--75835, 2026.

\bibitem[Zhang et~al.(2025{\natexlab{a}})Zhang, Li, Cheng, Hu, Yuan, Chen,
  Leng, Jiang, Zhang, Li, et~al.]{zhang2025videollama}
Boqiang Zhang, Kehan Li, Zesen Cheng, Zhiqiang Hu, Yuqian Yuan, Guanzheng Chen,
  Sicong Leng, Yuming Jiang, Hang Zhang, Xin Li, et~al.
\newblock {VideoLLaMA} 3: Frontier multimodal foundation models for image and
  video understanding.
\newblock \emph{arXiv preprint arXiv:2501.13106}, 2025{\natexlab{a}}.

\bibitem[Zhang et~al.(2024{\natexlab{a}})Zhang, Wang, Tang, Liu, Feng, Dai, and
  Jin]{zhang2024flash}
Haoji Zhang, Yiqin Wang, Yansong Tang, Yong Liu, Jiashi Feng, Jifeng Dai, and
  Xiaojie Jin.
\newblock {Flash-VStream}: Memory-based real-time understanding for long video
  streams.
\newblock \emph{arXiv preprint arXiv:2406.08085}, 2024{\natexlab{a}}.

\bibitem[Zhang et~al.(2024{\natexlab{b}})Zhang, Zhang, Li, Zeng, Yang, Zhang,
  Wang, Tan, Li, and Liu]{zhang2024long}
Peiyuan Zhang, Kaichen Zhang, Bo Li, Guangtao Zeng, Jingkang Yang, Yuanhan
  Zhang, Ziyue Wang, Haoran Tan, Chunyuan Li, and Ziwei Liu.
\newblock Long context transfer from language to vision.
\newblock \emph{arXiv preprint arXiv:2406.16852}, 2024{\natexlab{b}}.

\bibitem[Zhang et~al.(2025{\natexlab{b}})Zhang, Dong, Lin, Madotto, Kumar,
  Damavandi, Chai, and Moon]{zhang2025proactive}
Yichi Zhang, Xin~Luna Dong, Zhaojiang Lin, Andrea Madotto, Anuj Kumar, Babak
  Damavandi, Joyce Chai, and Seungwhan Moon.
\newblock Proactive assistant dialogue generation from streaming egocentric
  videos.
\newblock In \emph{Proceedings of the 2025 Conference on Empirical Methods in
  Natural Language Processing}, pages 12055--12079, 2025{\natexlab{b}}.

\bibitem[Zhang et~al.(2025{\natexlab{c}})Zhang, Shi, Wang, and
  Yang]{zhang2025eyes}
Yulin Zhang, Cheng Shi, Yang Wang, and Sibei Yang.
\newblock Eyes wide open: Ego proactive {Video-LLM} for streaming video.
\newblock \emph{arXiv preprint arXiv:2510.14560}, 2025{\natexlab{c}}.

\bibitem[Zheng et~al.(2026)Zheng, Ding, Yang, Jiang, Wu, Zhang, Wang, Cao, and
  Liu]{zheng2026garde}
Yikai Zheng, Xin Ding, Yifan Yang, Shiqi Jiang, Hao Wu, Qianxi Zhang, Weijun
  Wang, Ting Cao, and Yunxin Liu.
\newblock {Em-Garde}: A propose-match framework for proactive streaming video
  understanding.
\newblock In \emph{European Conference on Computer Vision}, pages 504--521.
  Springer, 2026.

\end{thebibliography}

}
\end{document}